\documentclass{article}
\usepackage{spconf,amsmath,graphicx}
\usepackage{amsfonts}       
\usepackage[flushleft]{threeparttable}
\usepackage{booktabs}       
\usepackage[noadjust]{cite}

\title{Row Conditional-TGAN for Generating Synthetic Relational Databases}
\name{Mohamed Gueye, Yazid Attabi, Maxime Dumas 
}
\address{Croesus Lab \\
	Croesus \\
	Laval, Québec, Canada \\}
\begin{document}
%
\maketitle
\begin{abstract}
Besides reproducing tabular data properties of standalone tables, synthetic relational databases also require modeling the relationships between related tables. In this paper, we propose the Row Conditional–Tabular Generative Adversarial Network (RC-TGAN), a novel generative adversarial network (GAN) model that extends the tabular GAN to support modeling and synthesizing relational databases. The RC-TGAN models relationship information between tables by incorporating conditional data of parent rows into the design of the child table's GAN. We further extend the RC-TGAN to model the influence that grandparent table rows may have on their grandchild rows, in order to prevent the loss of this connection when the rows of the parent table fail to transfer this relationship information. The experimental results, using eight real relational databases, show significant improvements in the quality of the synthesized relational databases when compared to the benchmark system, demonstrating the effectiveness of the RC-TGAN in preserving relationships between tables of the original database.
\end{abstract}
\section{Introduction}
\label{sec:intro}

Synthetic data generation is an active research topic in machine learning and aims to address data sparsity in development environments and in building data-driven models, especially when the data conveys private and sensitive information about individuals. In recent years, significant progress has been made in synthetic data generation models, particularly with the introduction of Generative Adversarial Networks (GANs) \cite{goodfellow_generative_2014}. While the performance of GANs is well established for a variety of tasks such as face \cite{brock_large_2018}, speech \cite{yu_seqgan_2017}, and open-domain dialogue \cite{li_adversarial_2017} generation, image \cite{zhu_unpaired_2017} and text-to-image \cite{zhang_stackgan_2016} translation, the generation of synthetic relational databases remains an under-investigated task, particularly within a deep learning approach.

In a relational database, there are multiple tables representing different collections of objects that may be interconnected. A single table represents a particular set of objects, arranged such that every column represents an attribute of the object. Each row represents an instance of the object, usually identified by a unique reference ID stored in the column known as the primary key of the table. To express a relationship, tables have columns containing a primary key ID of another table, i.e., the foreign key. The table with the primary key represents the parent table, while the table with the foreign key references represents the child table. A table may be a parent in one relationship and a child in another relationship.

Promising results are obtained with Tabular GANs (TGANs)
\cite{park_data_2018,chen_faketables_2019,xu_synthesizing_2018,xu_modeling_2019,kuo_generative_2020}, which are new versions of generative models designed to model and synthesize tabular data. Because of the complexity of tabular data, the scope of previous work was limited to tabular data in a single table with independent rows. 
Building a generative model to synthesize relational datasets is not a simple task. In addition to the challenges inherent to tabular data, such as modeling the distribution of each column and the relationships between the columns within the table, synthesizing relational databases requires modeling the relationships between tables in the same database, i.e., the relationship between parent and child tables. Indeed, generative models must support four types of parent-child relationships \cite{montanez_sdv_2018}: linear (table with only parent-child relationships), multiple-child (table with a primary key that is referenced by various tables), multiple-parent (table with columns that reference the primary keys of multiple tables), and multiple-child and multiple-parent. These schemas introduce new constraints that generative models must address. 

Research on synthetic relational databases has focused on statistical modeling. One state-of-the-art system is the Synthetic Data Vault (SDV) \cite{patki_synthetic_2016}. SDV builds sequentially generative models for individual tables and performs additional computations to account for the relationships between the tables. For each table, SDV computes aggregate statistics for the child tables that reference the current table. The aggregate statistics (i.e., the probability distributions of each column and the covariance matrices) are then added to the parent table, forming an extended table. Subsequently, a Gaussian Copula process is used to create a generative model of the extended table. The calculated distributions and covariances are used to sample new data.  
Overall, SDV successfully models a variety of relational datasets and uses the obtained generative models to synthesize new data. Specifically, SDV can easily address linear relationships in datasets; however, multiple-parent relationships are the most difficult to address. In this case, given that SDV generates a separate model for the child table in each parent's extended table, it becomes unclear which parameters should be used for data synthesizing. 
To the best of our knowledge, there is no publication on the generation of synthetic relational datasets based on GANs.

The objective of this paper is to extend the power modeling of GANs (CT-GAN in particular) to generate synthetic relational datasets (multi-tables), in order to better model the more complex column data distributions and intra- and inter-table correlations that conventional statistical methods have difficulty handling accurately, and to conveniently cover all relationship schemas between parent-child tables.
We introduce a new GAN model, the Row Conditional–Tabular GAN (RC-TGAN), which extends the original TGAN model to support relational datasets by incorporating conditional data from parent rows into the design of the GAN model corresponding to the child table. RC-TGAN has the inherent ability to address all relationship schemas without additional processing steps. 
We also extend RC-TGAN to maximize the capture of the influence that grandparent (or higher-level ancestor) rows may also have on their grandchild rows, thus preventing the loss of this connection when the parent table rows fail to transfer this relationship information.

This paper is organized as follows: Section II provides a background on tabular GANs for single table generation; Section III describes the proposed RC-TGAN model for relational database modeling and synthesis; Section IV reports on the experimental quality evaluation of the proposed model using various real and complex relational databases; finally, Section V concludes the paper.

\section{GANs for Tabular Data}
\label{sec:GAN}
Before describing the RC-TGAN model in more detail, we provide a brief overview of tabular GANs and define some of the notation used in the rest of the paper.
\subsection{Notation}
Let $\mathcal{D}$ be a relational database with $K$ tables $T^{(1)},\ldots,T^{(K)}$, and a table $T^{(i)}$ is defined by $p_{i}$ columns $X^{(i)}_{1},\ldots,X^{(i)}_{p_{i}}$ and a set of rows. The $k$-th row of the table $T^{(i)}$ can be considered a feature vector of $p_{i}$ dimensions $\left(x^{(i)}_{1,k},\ldots,x^{(i)}_{p_{i},k}\right)$. 
We also define $\mathcal{P}_{a}\left(T\right)$ as the set of parents of table $T$.

\subsection{Tabular GANs}
GAN is a deep learning-based generative model designed to map noisy sample vectors $z$ from a prior distribution $\mathbb{P}_{z}$ into samples that have data distribution similar to the real data distribution $\mathbb{P}_{r}$. GANs are implemented using two neural networks with an antagonist objective: \textit{Generator} $(G)$  and \textit{Discriminator} $(D)$. $G$ tries to generate realistic data to trick the discriminator. $D$ is a binary classifier that tries to distinguish fake data (made up by $G$) from real data. Both models $G$ and $D$ are trained together in a zero-sum game, adversarial, until $G$ generates satisfying samples. The described adversarial training can be formulated as the following minimax problem:
\begin{multline}
\min_{G}\max_{D} V(D, G)=\mathbb{E}_{x\sim \mathbb{P}_{r}}\left[\log D(x)\right]\\+\mathbb{E}_{z\sim \mathbb{P}_{z}}\left[\log (1-D(G(z)))\right],
\end{multline}
where $V(D, \theta)$ denotes the value function, $x$ is the feature vector from the real data distribution, $D(x)$ and $G(x)$ are the outputs of $D$ and $G$, and $\mathbb{E}$ denotes the expected value. 
The \textit{Wasserstein} GAN (WGAN)\cite{gulrajani_improved_2017} is a GAN variant whose discriminator is called the \textit{critic}. The critic $(C)$ inputs fake or real data and outputs a score instead of binary output.  

Several modifications have been introduced to adapt the classical GAN to the specificities of tabular datasets, such as addressing mixed-type columns and highly imbalanced categorical columns. FakeGAN\cite{chen_faketables_2019}, TGAN\cite{xu_synthesizing_2018}, and Conditional TGAN (CTGAN)\cite{xu_modeling_2019} are examples of such GAN variants.  
CTGAN is based on WGAN and allows modeling a tabular data distribution and sampling new rows of a given single table. CTGAN differs from TGAN in how it handles discrete columns. CTGANs train TGANs conditionally by randomly selecting a category of discrete columns in each iteration. CTGANs are particularly effective in modeling tables with significant categorical column imbalance.

Because of their favorable properties and performance, we aim to extend the capabilities of CTGAN to model relational databases and synthesize new databases while preserving the relationship information interconnecting tables.

\section{Description of the Proposed RC-TGAN Model}
\label{section:desc_gan}

\begin{figure}[htbp]
	\centering
	\includegraphics[scale=0.45] {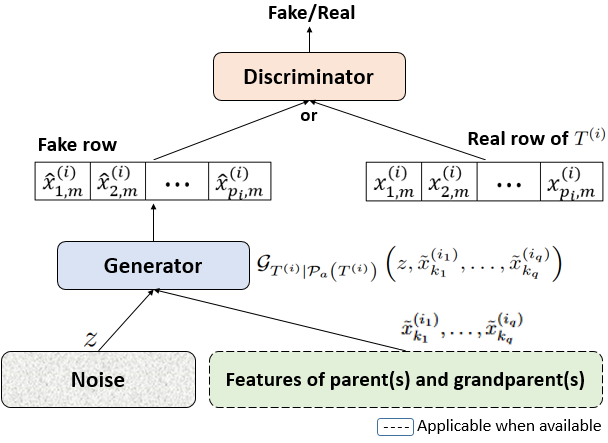}
	\caption{RC-TGAN for single table modeling}
	\label{fig:fig1}
\end{figure}

This section describes the design of the RC-TGAN model, depicted in Fig. \ref{fig:fig1}. The purpose of the RC-TGAN is to model any relational database and to synthesize new databases that can replace or extend the original databases. 
\subsection{Database Modeling}
The relational database $\mathcal{D}$ follows an unknown distribution $\mathbb{P}_{\mathcal{D}}$. Because of the complexity of relational databases, the distribution $\mathbb{P}_{\mathcal{D}}$ is difficult to tackle directly. We therefore proceeded by modeling table by table, while taking into account the parent-child relationships. Previous work related to modeling single tables using tabular GANs assumed that each row of a table $T^{(i)}$ is generated by a distribution $\mathbb{P}_{T^{(i)}}$ on its features space  $X^{(i)}_{1},\ldots,X^{(i)}_{p_{i}}$ independently of other tables:
\begin{equation} \label{fig:true_dist_without_par}
\mathbb{P}_{T^{(i)}}\left(\cdot \right) \sim \tilde{x}_{k}^{(i)} \equiv \left(x_{1,k}^{(i)},\ldots,x_{p_{i},k}^{(i)}\right)
\end{equation}
In this paper, we assume that the relationship information linking the tables can be extracted from the rows of the parent tables when modeling or generating the child table rows. Specifically, each row of a table $T^{(i)}$ with $q$ parent tables $\mathcal{P}_{a}\left(T^{(i)}\right) = \left\{T^{(i_{1})},\ldots,T^{(i_{q})}\right\}$, is generated by a conditional distribution $\mathbb{P}_{T^{(i)}|{P}_{a}\left(T^{(i)}\right)}$ given the features of its rows parents:
\begin{multline} \label{fig:true_dist_with_par}
\mathbb{P}_{T^{(i)} | {P}_{a}\left(T^{(i)}\right)}\left(\cdot \vert \tilde{x}_{k_{1}}^{(i_{1})},\ldots,\tilde{x}_{k_{q}}^{(i_{q})} \right) \sim \tilde{x}_{k}^{(i)} 
\end{multline}
where $\tilde{x}_{k_{w}}^{(i_{w})} \equiv \left(x_{1,k_{w}}^{(i_{w})},\ldots,x_{p_{i_{w}},k_{w}}^{(i_{w})}\right)$ represents the feature vector of the $k_w$-th row of the $i_w$-th parent table, for $ w \in \left\{1,\ldots,q\right\}$. This means that each parent row affects the distribution of the child table rows.
$\mathcal{G}_{T^{(i)}|\mathcal{P}_{a}\left(T^{(i)}\right)}$ – the generator of RC-TGAN synthesizing data from table $T^{(i)}$ – takes as input the noise vector $z$ in addition to the features of the parent rows. Furthermore, $\mathcal{G}_{T^{(i)}|\mathcal{P}_{a}\left(T^{(i)}\right)}\left(z,\tilde{x}^{(i_{1})}_{k_{1}},\ldots,\tilde{x}^{(i_{q})}_{k_{q}} \right)$ represents a mapping to row space $T^{(i)}$.

Note that the distribution of root tables in a database, i.e., tables with no parent(s), are modeled without inputs from other tables, as in conventional tabular GANs, and the synthetic data for a table $T^{(i)}$ are generated using a generator $\mathcal{G}_{T^{(i)}}$. In this context, the tabular GAN is considered a special case of the general RC-TGAN model. 

\subsection{Database Synthesis}
Data synthesis is based on the row conditional generator of RC-TGAN trained for each table. In contrast to the modeling step, the order in which the tables are synthesized must adhere to the following rule: All the parent tables $\mathcal{P}_{a}\left(T^{(i)}\right)$ must be synthesized before the child table $T^{(i)}$, i.e., we must first sample the table with no parent(s), and then sample the tables for which parents are already sampled until we cover all the tables in the database. This allows using the synthesized parent rows as features when generating child table rows to create the relationship information.
This rule is applicable if the collection of parent-child relationships is acyclic, i.e., no table is an ancestor of itself.

\subsection{Extended RC-TGAN}
Besides modeling the relationship between the child and its immediate parent tables, we seek to preserve  the information linking a child table with its grandparent tables (and its ancestor tables in general) in the synthetic data generated by RC-TGAN. This is based on the assumption that the relationship information between the child and grandparent rows is not always preserved and transmitted via the parent table row. This is particularly true when the parent table has a limited set of features (i.e., the remaining columns, after excluding primary and foreign keys). This limited set of features is unable to convey the required information to model the desired relationship. By including ancestor features, we are able to preserve more information about the relationship and therefore improve the quality of the generated relational synthetic data. This variant of RC-TGAN can be applied to tables in relational databases with grandparent or higher level ancestor tables.
Thus, we expand the set of features used as conditional data of the child table in RC-TGAN to include ancestor features when this information is available, instead of limiting ourselves to parent features.

\section{Experimental Results}
\label{sec:others}
This section describes the experimental setup we used to validate the fidelity of the data generated using the RC-TGAN compared to real data.
To show the effectiveness of the proposed model, we use SDV as a benchmark system for performance comparison. The steps of the experimental process consist of building the RC-TGAN model for each table, generating new synthetic data using the created RC-TGAN, and then evaluating the synthesized data using the evaluation metrics. In the following sections, we will first describe the experimental methodology, and then present the results with a discussion.

\subsection{Relational Datasets}
\label{ssec:rdb}

We extensively evaluated RC-TGAN using the following eight public databases that encompass the different and complex relationship schemas: \textit{Airbnb}\cite{noauthor_kaggle_nodate}, \textit{Rossmann}\cite{noauthor_rossmann_nodate}, \textit{Biodegradability}\cite{blockeel_experiments_2004}, \textit{Mutagenesis}\cite{debnath_structure-activity_1991} (representing datasets with linear relationships), \textit{Telstra}\cite{noauthor_telstra_nodate}, \textit{Walmart}\cite{tommy_wilczek_walmart_nodate} (for multiple-child relationships), \textit{World Development Indicators}\cite{noauthor_world_nodate}, and \textit{Coupon Purchase Prediction}\cite{noauthor_coupon_nodate} (for multiple child and parent relationships).
 
\subsection{Experimental Setup}
\label{sssec:expSetup}

The quality of the synthetic relational databases sampled by RC-TGAN are evaluated using detection metrics. Detection metrics assess the difficulty of distinguishing synthetic data from real data by using a binary machine learning classifier trained on the two labeled classes: real data and synthetic data.
We use the logistic model as a binary classifier to evaluate two performance aspects: (i) the quality of the standalone synthetic tables by evaluating each table independently, referred to as \textit{logistic detection} metric (LD); (ii) the ability to preserve the parent-child relationship by applying the logistic detection classifier on the denormalized synthetic tables, referred to as \textit{parent-child logistic detection} metric (P-C LD).
LD and P-CLD scores, implemented in the SDV toolkit \cite{patki_synthetic_2016}, are computed as follows:
\begin{multline} \label{equat:LD_metric}
score = 1-[2\times\max(0.5,au\_roc)-1],
\end{multline} 
where $au\_roc$ represents the \textit{area under} the \textit{receiver operating characteristic} (ROC) \textit{curve} achieved by the classifier used for LD or P-CLD. \textit{Score} values range from 0 to 1, and higher \textit{score} values indicate better quality of the synthesized data, i.e., the classifier cannot distinguish real from synthetic rows. 

RC-TGAN1 (based on conditional data of the parent row) and RC-TGAN2 (extended to grandparent rows) are implemented by modifying the source code of CTGAN, packaged in the SDV library. The hyperparamater tuning of the RC-TGAN neural networks is carried out separately for each single table. 

\subsection{Results and Discussion}
\label{sec:results}

Table \ref{table:table_result} shows the results of LD and P-C LD for SDV, RC-TGAN1, and RC-TGAN2, when applicable to the eight databases. 
First, we observe that significant improvements in both evaluation metrics are achieved using the two versions of RC-TGAN on most databases when compared to SDV. On average, relative improvements of 14.42\% and 11.68\% are obtained for LD and P-CLD respectively for RC-TGAN-1, and 16.53\% and 24.35\% are obtained for LD and P-CLD respectively for RC-TGAN-2.

Second, the P-CLD results for RC-TGAN-2 in Table \ref{table:table_result} show that incorporating grandparent information better preserves the relationship information between tables in the relational database compared to using the parent information alone (RC-TGAN-1). Interestingly, we observe that RC-TGAN-2 also outperforms RC-TGAN-1 on the LD metric for both databases, Biodegradability and Mutagenesis. These results suggest that, besides the relationship information between tables, grandparent information also contributes to preserving the shape of the data distribution and/or the correlation information between columns within the same table of the original data. Finally, we note that the results of P-CLD are significantly lower compared to LD scores, which demonstrates the challenging task of modeling and synthesizing relational databases that preserve the complete relationship information between tables.

\begin{table}[h] 
	\centering
	\caption{LD and P-C LD results for SDV, RC-TGAN1 and RC-TGAN2. * indicate the relational databases with grandparent tables that can benefit from the RC-TGAN-2 model.} 
	\begin{threeparttable}
		\begin{tabular}{ll p{1.4cm} p{1.4cm} p{1.4cm}}
			\multicolumn{1}{l}{\textbf{Datasets}} & \textbf{Metrics} &  \textbf{\centering SDV} & {\centering \textbf{RC-TGAN-1}} & {\centering \textbf{RC-TGAN-2}} \\
			\specialrule{1.5pt}{0.75pt}{0.75pt}
			\multicolumn{1}{l}{Rossman} & LD & 62.67\% & \textbf{72.06\%} & \textbf{72.06\%} \\
			& P-C LD & 44.44\% & \textbf{56.52\%} & \textbf{56.52\%} \\
			\multicolumn{1}{l}{Telstra} & LD & 82.80\% & \textbf{95.93\%} & \textbf{95.93\%} \\
			& P-C LD & 75.52\% & \textbf{88.08\%} & \textbf{88.08\%} \\
			\multicolumn{1}{l}{Walmart} & LD & 33.34\% & \textbf{46.79\%} & \textbf{46.79\%} \\
			& P-C LD & 00.03\% & \textbf{15.67\%} & \textbf{15.67\%} \\
			\multicolumn{1}{l}{World-} & LD & 31.90\% & \textbf{42.78\%} & \textbf{42.78\%} \\
			indicators & P-C LD & 19.22\% & \textbf{26.26\%} & \textbf{26.26\%} \\
			\multicolumn{1}{l}{Airbnb} & LD & 50.03\% & \textbf{51.62\%} & \textbf{51.62\%} \\
			& P-C LD & \textbf{08.63\%} & 06.50\% & 06.50\% \\
			\multicolumn{1}{l}{Coupon} & LD & 41.95\% & \textbf{50.00\%} & \textbf{50.00\%} \\
			& P-C LD & 19.06\% & \textbf{20.50\%} & \textbf{20.50\% }\\  
			\midrule 
			\multicolumn{1}{l}{Biodegra- } & LD & 45.83\% & 47.20\% & \textbf{49.30\%} \\
			dability\tnote{*} & P-C LD & 27.45\% & 08.92\% & \textbf{31.74\%} \\
			\multicolumn{1}{l}{Mutagenesis \tnote{*}} & LD & 42.30\% & 40.80\% & \textbf{46.93\%} \\
			& P-C LD & 26.50\% & 24.20\% & \textbf{29.36\%} \\
			\specialrule{1.5pt}{0.75pt}{0.75pt}
			\multicolumn{1}{l} {\textbf{Average}} & LD & 48.85\% & 55.90\% & \textbf{56.93\%} \\
			& P-C LD & 27.61\% & 30.83\% & \textbf{34.33\%} \\
		\end{tabular}%
	\end{threeparttable}
	\label{table:table_result}
\end{table}%
\section{Conclusion}
\label{sec:concl}
In this paper, we presented RC-TGAN, a novel tabular GAN for modeling and synthesizing relational databases. RC-TGAN models the relationship information between tables by incorporating conditional data on parent rows into the GAN model child table. The purpose of the RC-TGAN is therefore to reproduce the effect of the parent rows on the distribution of the child rows. On average, relative improvements of 14.42\% and 11.68\% were achieved for LD and P-CLD metrics respectively, obtained on eight real relational databases when compared to the SDV model. We also found that extending conditional data to grandchild rows in RC-TGAN, when available, allowed acquiring and modeling additional relationship information that parent tables may fail to transfer to child tables, specifically when the parents have a limited set of attributes. Incorporating grandparent information resulted in relative improvements of 16.53\% and 24.35\% for LD and P-CLD respectively. Finally, a significant gap still exists between the P-CLD and LD results, which demonstrates the challenging task of preserving relationship information between tables. Reducing this gap will be addressed in future work.
\vfill\pagebreak
\newcommand{\BIBdecl}{\setlength{\itemsep}{-0.01 em}}
\bibliographystyle{IEEEbib}
\bibliography{refs_rctgan}  
\end{document}